\documentclass[letterpaper]{article} 
\usepackage{aaai2026}  
\usepackage{times}  
\usepackage{helvet}  
\usepackage{courier}  
\usepackage[hyphens]{url}  
\usepackage{graphicx} 
\usepackage{natbib}  
\usepackage{caption} 
\usepackage{enumitem}
\usepackage{bm}
\usepackage{booktabs}
\usepackage{amsfonts}
\usepackage{amsmath}
\usepackage{adjustbox}
\usepackage{amsthm}
\usepackage{multirow}
\usepackage[ruled,vlined,linesnumbered]{algorithm2e}
\usepackage{algorithmic}
\usepackage{xcolor}
\usepackage[titletoc]{appendix}
\usepackage{newfloat}
\usepackage{listings}
\usepackage{float}
\DeclareCaptionStyle{ruled}{labelfont=normalfont,labelsep=colon,strut=off} 
\title{Neural Graph Navigation for Intelligent Subgraph Matching}
\author{
    Yuchen Ying\textsuperscript{\rm 1,3}\equalcontrib,
    Yiyang Dai\textsuperscript{\rm 1,3}\equalcontrib, 
    Wenda Li\textsuperscript{\rm 1}\equalcontrib, 
    Wenjie Huang\textsuperscript{\rm 1,3}, 
    Rui Wang\textsuperscript{\rm 1,3}, \\
    Tongya Zheng\textsuperscript{\rm 1,2,3}\corresponding, 
    Yu Wang\textsuperscript{\rm 1,3}, 
    Hanyang Yuan\textsuperscript{\rm 1,3}, 
    Mingli Song\textsuperscript{\rm 1,3}
}
\affiliations{
    \textsuperscript{\rm 1}State Key Laboratory of Blockchain and Data Security, Zhejiang University\\
    \textsuperscript{\rm 2}Zhejiang Provincial Engineering Research Center for Real-Time SmartTech in Urban Security Governance, School of Computer and Computing Science, Hangzhou City University\\
    \textsuperscript{\rm 3}Hangzhou High-Tech Zone (Binjiang) Institute of Blockchain and Data Security\\
    
    \{yingyc, yiyangdai, lwdup, wjie, rwang21\}@zju.edu.cn\\
    doujiang\_zheng@163.com, \{yu.wang, yuanhanyang, brooksong\}@zju.edu.cn

}

\usepackage{bibentry}

\begin{document}

\maketitle

\begin{abstract}
Subgraph matching, a cornerstone of relational pattern detection in domains ranging from biochemical systems to social network analysis, faces significant computational challenges due to the dramatically growing search space. Existing methods address this problem within a filtering-ordering-enumeration framework, in which the enumeration stage recursively matches the query graph against the candidate subgraphs of the data graph. However, the lack of awareness of subgraph structural patterns leads to a costly brute-force enumeration, thereby critically motivating the need for intelligent navigation in subgraph matching. To address this challenge, we propose Neural Graph Navigation (NeuGN), a neuro-heuristic framework that transforms brute-force enumeration into neural-guided search by integrating neural navigation mechanisms into the core enumeration process. By preserving heuristic-based completeness guarantees while incorporating neural intelligence, NeuGN significantly reduces the \textit{First Match Steps} by up to 98.2\% compared to state-of-the-art methods across six real-world datasets. 
\end{abstract}

\section{Introduction}

Subgraph matching is a cornerstone of graph-based analysis, enabling the precise identification of query patterns within large-scale networks. It underpins a wide array of critical applications: discovering evolutionarily conserved motifs in protein interaction networks~\cite{jeong2001protein1, saha2017protein_motifs, grindley1993identification}, detecting anomalous behavior in social platforms~\cite{ rehman2020online, wang2024spatiotemporal}, and supporting semantic analysis in knowledge-driven question answering~\cite{hu2017answering}. By finding isomorphic instances of a query subgraph in a larger data graph, subgraph matching provides a principled mechanism for structural reasoning across domains.

 The prohibitive computational cost of subgraph matching poses significant challenges in real-world applications. In the general case, subgraph matching is NP-hard~\cite{hartmanis1982computers}, with worst-case time complexity $O(|V_G|^{|V_Q|})$, where $|V_G|$ and $|V_Q|$ denote the number of vertices in the data graph $G$ and the query graph $Q$, respectively. To mitigate this challenge, most state-of-the-art methods adopt a \textit{filtering-ordering-enumeration} framework that systematically reduces complexity across phases. The filtering phase prunes unpromising nodes, the ordering phase determines the matching sequence of the query graph, and the enumeration phase exhaustively identifies valid matches. However, the lack of subgraph structural pattern awareness during the enumeration phase results in a \textbf{brute-force enumeration} procedure. 
 This drawback motivates the need for \textbf{intelligent navigation} in the enumeration phase to adaptively prioritize the candidate nodes during subgraph matching.

 Recent advances in graph representation learning~\cite{kipf2016gcn,dwivedi2020gt2,zhang2025effective} suggest that Graph Neural Networks (GNNs) can capture rich structural patterns~\cite{xu2024temporal, wang2024spatiotemporal2}, demonstrating significant potential for application in subgraph matching. 
Nonetheless, GNNs designed for feedforward predictions are unable to perform search and backtracking directly in subgraph matching. Further, Neural algorithmic reasoning~\cite{velivckovic2019neural_graph_algo,velivckovic2022clrs} design a learning framework for GNNs to simulate combinatorial algorithms; however, it is incapable of scaling to the exponentially growing search space of subgraph matching. 
NeuroMatch~\cite{lou2020neural} predicts the existence of subgraphs, while cannot locate concrete matches. 
Beyond these end-to-end methods, hybrid approaches integrate GNNs into classical frameworks. 
 Pruning-based methods~\cite{duong2021efficient,ye2024GNNPE} utilize GNNs for candidates pruning, but may overlook potential matches due to the probabilistic uncertainty of neural networks.
In contrast, reinforcement learning methods~\cite{wang2022RLQVO,li2025rsm} optimize the matching order of the query  without changing the burdensome enumeration core. 
Consequently, the final enumeration phase still relies on brute-force enumeration without structure-aware intelligent navigation.

However, integrating navigation into the matching process remains two technical challenges. 
First, aligning the query graph with the evolving structures of partial matches and providing navigation informed by the global perception of the data graph is a non-trivial problem.
Second, subgraph matching requires a specialized training objective, as traditional graph pretraining tasks do not capture the structural correspondence and search dynamics inherent in matching. 

To address these challenges, we design a \textit{neuro-heuristics fusion} mechanism to navigate the subgraph matching algorithm using prioritized enumeration orders while preserving enumeration completeness. 
The proposed Neural Graph Navigation (NeuGN) serves as a plug-and-play framework that transforms subgraph enumeration into a neural generative process. 
NeuGN integrates two synergistic components: a \textbf{Query Structure Extractor (QSExtractor)} that compresses query graph structural patterns into latent navigation signals using GNNs, and a \textbf{Generative Graph Navigator (GGNavigator)} that formulates search path construction as a sequential cloze generation task~\cite{sun2019bert4rec}, leveraging a Transformer architecture to progressively replace padding tokens with predicted node identifiers. At each enumeration step, the navigator injects structural awareness by maintaining a candidate queue ranked by node matching confidence. For practical deployment, our batched inference strategy parallelizes neural evaluations across search branches. To the best of our knowledge, we presents the first framework that integrates generative neural navigation into the enumeration phase for subgraph matching, guaranteeing completeness while significantly reducing the \textit{First Match Steps} through neural intelligence.

Our contributions are summarized as follows:
\begin{itemize}
 \item To the best of our knowledge, this is the first work to integrate learned navigation directly into the core enumeration phase of subgraph matching. 
 
\item We propose a novel plug-and-play Neural Graph Navigation Framework, unifying query-aware structural perception with global structure-aware generative navigation.

\item Extensive experiments on six real-world datasets show that NeuGN reduces \textit{First Match Steps} by up to 98.2\% compared to state-of-the-art methods.

\end{itemize}
\section{Related Work}
\textbf{Traditional Subgraph Matching.} 
As summarized in a subgraph matching survey~\cite{zhang2024survey}, most traditional algorithms follow the filtering-ordering-enumeration framework. Early algorithms like Ullmann’s~\cite{ullmann1976algorithm} introduced backtracking-based search, later refined by VF2~\cite{vf2} with advanced pruning rules. Existing subgraph matching algorithms can be broadly categorized into two groups. The first group~\cite{bhattarai2019ceci,han2019dpiso,han2013turboiso,kim2021veq,rivero2017sgmatch,shang2008taming,sun2020memory,sun2020subgraph} employs precomputed indices or auxiliary data structures to prune irrelevant candidates prior to enumeration. The second group~\cite{arai2023gup,jin2023circinus,li2024newsp,ive,bice} focuses on developing pruning strategies that operate dynamically during the subgraph enumeration process, without relying on auxiliary structures. \\
\textbf{Learning Enhanced Subgraph Matching.} 
With the rapid development of machine learning, an increasing number of researchers have begun to explore its potential to enhance subgraph matching. NeuroMatch~\cite{lou2020neural} and AEDNet~\cite{lan2023aednet} remove the backtracking search entirely, relying solely on graph-level embeddings to predict subgraph existence, which cannot locate concrete matches. Pruning-based methods~\cite{duong2021efficient, ye2024GNNPE, yang2025gnnae} deploy GNNs as pruning tools, but may omit potential matches due to the probabilistic nature of neural networks.
Moreover, recent studies such as RLQVO~\cite{wang2022RLQVO} and RSM~\cite{li2025rsm} have shown that incorporating reinforcement learning frameworks can effectively optimize the reordering phase.  
However, the aforementioned approaches still depend on the brute-force enumeration process, highlighting the need for intelligent perception-driven navigation.

\section{Preliminaries}
\noindent
In this study, we concentrate on undirected connected graphs where nodes are labeled. 
Let $Q = (V_Q, E_Q, \Sigma, L_Q)$ be a query graph and $G = (V_G, E_G, \Sigma, L_G)$ be a data graph, where $V_Q$ and $V_G$ are the vertice sets, $ E_Q $ and $ E_G $ are the edge set, $ \Sigma $ denotes the label set, $ L_Q: V_Q \rightarrow \Sigma $ and $ L_G: V_G \rightarrow \Sigma $ map each node to its respective label.
For convenience, let $ d(u) $ be node $u$'s degree, and $ N_Q(u) $ and $ N_G(u) $ denote the neighbor sets of $u$ in the query graph and data graph, respectively.

\textbf{Subgraph Matching.} Given a \textit{query graph} $Q = (V_Q, E_Q, \Sigma, L_Q)$ and a \textit{data graph} $G = (V_G, E_G, \Sigma, L_G)$, we say that $Q$ is subgraph isomorphic to $G$ if there exists a mapping function $f:V_Q \rightarrow V_G$ such that: \textit{(1)} $\forall u \in V_Q, \textit{ we have } L_Q(u) = L_G(f(u))\textit{ where } f(u) \in V_G,$ \textit{(2)}$\forall e(u_i, u_j) \in E_Q, \textit{ we have } e(f(u_i), f(u_j)) \in E_G, \textit{ and }$ \textit{(3)} $\forall u_i, u_j \in V_Q, u_i \neq u_j, \textit{ then } f(u_i) \neq f(u_j).$ The mapping function $f$ is also referred to as a \textit{match} (also called embedding) of $Q$ in $G$. Each match can be represented as a set of one-to-one node pairs $\{(u, f(u))\}$, where each node pair is termed an assignment. In the context of subgraph matching, the objective is to identify and return all possible matches of the query graph $Q$ within the data graph $G$, ensuring that each match satisfies the node and edge correspondence conditions specified above. \textit{Local candidate nodes} in $G$, generated in each enumeration parse, is a nodes list that satisfies the local constraints imposed by a specific node in $Q$, serving as a potential match in the mapping process.

\section{Method}
\begin{figure}[h]
    \centering 
    \includegraphics[width=0.95\columnwidth]{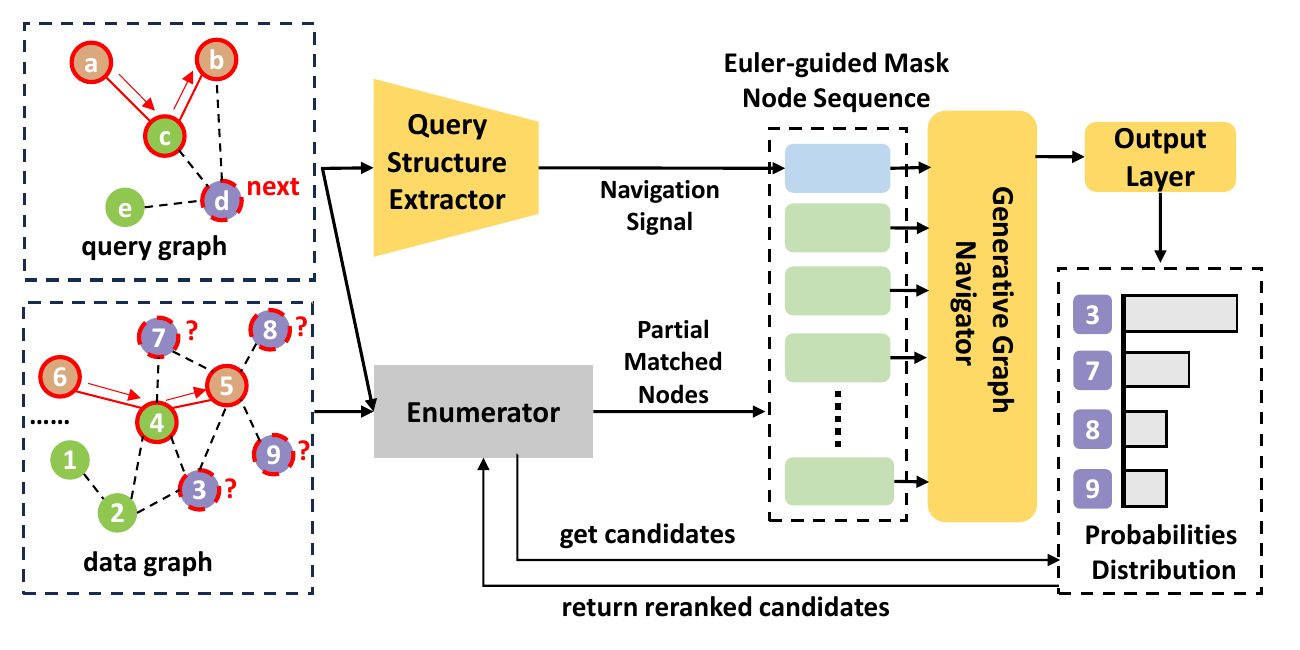}
    \caption{The illustrative diagram of NeuGN Framework. }
    \label{fig:Framework}
\end{figure}

Figure~\ref{fig:Framework} illustrates the overall framework of our proposed Neural Graph Navigation (NeuGN), seamlessly integrating traditional heuristic-based subgraph matching algorithms with neural-based intelligent navigation.
Following the traditional phases of filtering, ordering, and enumeration, NeuGN incorporates two neural modules into the algorithms: a Query Structure Extractor (QSExtractor) extracting query graph structures into latent navigation signals and a Generative Graph Navigator (GGNavigator) leveraging the signals to transform brute-force enumeration into a holistic structure-aware navigation process. 
This neural-enhanced strategy, where QSExtractor crystallizes the ``target navigation signal'' and GGNavigator plots the ``route'', constitutes NeuGN's core innovation.
The overall workflow is demonstrated in Algorithm~\ref{alg:subgraph} , where the critical parts are \textbf{highlighted}. More details are provided in Appendix B.1.

\begin{algorithm}[t]
\small
\caption{NeuGN Framework for Subgraph Matching.} 
\label{alg:subgraph}
\KwIn{A query graph \( Q \) and a data graph \( G \).} 
\KwOut{All the matches of \( Q \) into \( G \).}    

\BlankLine 
\( \bm{h_Q} \leftarrow \textbf{QSExtractor}\bm{(Q)} \;\) 

\( C \leftarrow \text{FilterNodes}(Q, G)\)\; 

\( \varphi \leftarrow \text{GenerateOrder}(Q, G, C)\)\;

\(\text{Enumerate}(Q, G, C, \varphi, \{ \}, \bm{h_Q}, 0)\)\; 

\BlankLine
\SetKwProg{Fn}{Function}{:}{} 
\Fn{\text{Enumerate}(Q, G, C, \(\varphi\), M, \(\bm{h_Q}\), i)}{ 
    \If{Termination condition met}{ 
        \text{output} \text{TerminateFunc}\((Q, G, C, \varphi, M)\)\;
    }
    \( u \leftarrow \text{SelectNextQueryNode}(Q, G, C, \varphi, M) \)\;
    
    \( C_M(u) \leftarrow \text{ComputeLocalCandidates}(Q, G, C, \varphi, M, i) \)\;
    
    \( \bm{Conf} \leftarrow \textbf{GGNavigator}\bm{(Q}, \bm{G}, \bm{C}, \bm{h_Q}, \bm{\varphi}, \bm{M)} \)  
    
    \( \bm{C_M(u)} \leftarrow \textbf{Sort}\bm{(C_M(u)}, \bm{Conf)} \)\;  
    
    \ForEach{\( v \in C_M(u) \)}{

        \(\text{PruningFunc}(Q, G, C, \varphi, M)\)\;
    
        Extend \( M \) by \( (u, v) \)\;
        
        \(\text{Enumerate}(Q, G, C, \varphi, M, i + 1)\)\; 
        
        Delete \( (u, v) \) from \( M \)\;
        
    }
}
\end{algorithm}

\subsection{Query Structure Extractor}
\label{subsec:Extractor}

Effective navigation of subgraph enumeration in NeuGN depends on a well-structured search path, which is informed by the structure patterns of the query graph. To address this need, the QSExtractor is designed to encode the structure of the query graph $Q$ into compact latent navigation signals. These signals capture the essential structural characteristics of $Q$ and serve as a structural compass for the downstream GGNavigator, guiding the search process effectively.

To encode node-label constraints vital for matching, we project discrete labels $L_Q$ into a semantic space:
\begin{equation}
    {Z}_Q = \text{Embed}(L_Q), \quad \text{Embed}: \mathbb{N}^l \to \mathbb{R}^{l \times d}.
\end{equation}
This embedding captures discriminative label semantics while modeling co-occurrence patterns and semantic correlations essential for matching.

To encode the query graph's structure information essential for navigation, we employ a Graph Convolutional Network (GCN) to propagate and distill structural dependencies. The layer-wise feature transformation is given by:
\begin{equation}
    H^{(l+1)} = \sigma\left( \hat{A} H^{(l)} W^{(l)} \right),
\end{equation}
where $\hat{A}$ is the normalized adjacency matrix with self-loops, $H^{(0)}$ is initialized from ${Z}_Q$, $W^{(l)}$ are learnable parameters and $\sigma(\cdot)$ is the ReLU activation function. 
The final node representations $\{h_v\}$ capture $Q$'s $L$-hop subgraphs.

To derive a navigation signal, we apply max-pooling over all node representations:
\begin{equation}
    h_Q = \text{MaxPool}(\{ h_v \mid v \in V_Q \}).
\end{equation} 
The output signal $h_Q$ is a compact representation of the structural pattern of $Q$, enabling the GGNavigator to compare the evolving substructures of $G$ with predefined targets.

\begin{figure*}[h]
    \centering 
    \includegraphics[width=0.95\linewidth]{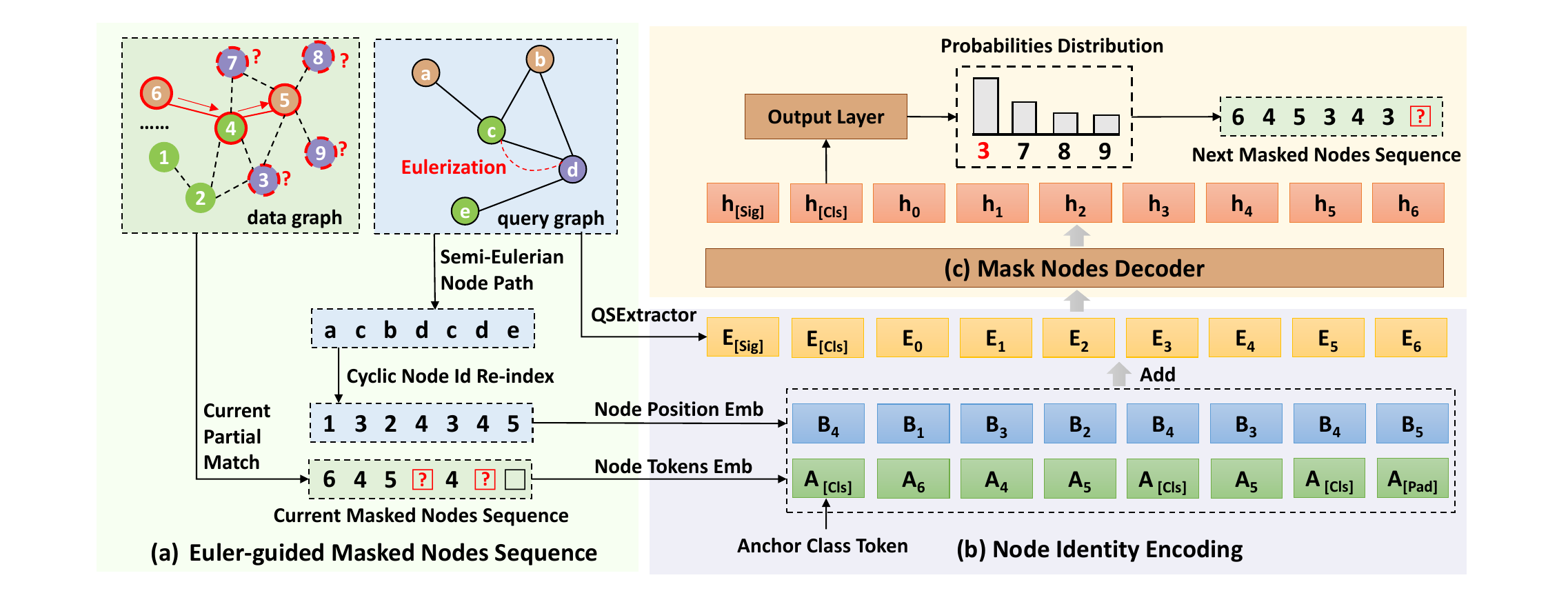}
 \caption{The illustrative diagram of Generative Graph Navigator.}
    \label{fig:decoder}
\end{figure*}

\subsection{Generative Graph Navigator}
To conquer the brute-force enumeration of subgraph matching, Generative Graph Navigator addresses three technical challenges: tracking dynamically evolving partial matches through \textit{Euler-Guided Masked Nodes Sequence}; maintaining persistent global awareness of the data graph via \textit{Node Identity Encoding} that anchors topological context; and
dynamically prioritizing high-likelihood candidates using confidence-based ranking of the \textit{Masked Nodes Decoder}.

\textbf{Euler-guided Masked Nodes Sequence.}
\label{subsec:masknode}
In NeuGN, we serialize the query graph into a Masked Nodes Sequence via Eulerian path transformation. This is achieved by duplicating edges to construct a (semi-)Eulerian path, which guarantees lossless graph reconstruction up to isomorphism by encoding all adjacency relationships as sequential dependencies~\cite{grohe2020graph}. Crucially, the  (semi-)Eulerian node path facilitates Transformer Decoder to implicitly learn topological constraints during generation, as adjacent tokens inherently capture edge connectivity. Consequently, the graph matching task is naturally transformed into a sequential cloze generation task. We follow the established method in~\cite{zhao2025graphgpt} to ensure consistency. 

Subsequently, we apply node position cyclic re-indexing to assign position IDs to each node occurrence in the (semi-)Eulerian path. Let $i \in \{0, 1, \dots, L-1\}$ denote the zero-based index ID, where $L$ is the path length. The cyclic re-indexing follows $i' = (i + r) \mod N$, where $r$ is a random offset and $N$ is a hyperparameter. This mitigates ordering bias during training, and crucially, the resulting position ID serves as a unique identity signature. By preserving the relative order of node appearances, these IDs inherently encode the structure of the subgraph, enabling the model to track connectivity patterns in the evolving partial match.

We illustrate the generation process of the Euler-guided Masked Nodes Sequence in Figure~\ref{fig:decoder}. First, we eulerize the query graph by adding auxiliary edges (e.g. duplicating edge $c \to d$ to form the path $a \to c \to b \to d \to c \to d \to e$). Assuming a random offset $r=1$, we assign node position IDs 1, 2, 3, 4, and 5 to nodes $a$, $b$, $c$, $d$, and $e$, respectively. The resulting node sequence is then tokenized into a Masked Nodes Sequence, where matched nodes are assigned their corresponding data graph IDs, unmatched positions are filled with padding tokens (\textcolor{black}{black box}), and the positions of next candidate node are marked with class tokens (\textcolor{black}{red box}) to guide the model toward the correct generation target at each step. 
As matching progresses, padding tokens are progressively replaced with actual matched node IDs, while the class token advances to indicate the next prediction position. More details can be found in Appendix B.

\textbf{Node Identity Encoding.}
\label{subsec:Embedding}
To holistically encode topological context while preventing structural information loss during graph serialization, we integrate two complementary embeddings into a unified representation. 
The Node Token Embedding matrix $A \in \mathbb{R}^{(|V_n|+2) \times d}$ preserves global node identities by mapping discrete node IDs (including padding and class tokens) to continuous vectors, with $|V_n|$ denoting the unique node count of the data graph. This anchors neighborhood connectivity patterns and structural semantics within the original graph. The Node Position Embedding matrix $B \in \mathbb{R}^{N \times d}$ encodes cyclically re-indexed positions from the Eulerian path to distinguish node sequencing roles, initialized orthogonally~\cite{kim2022pure} to prevent feature collapse. 
As illustrated in Figure~\ref{fig:decoder}, the framework processes the Masked Nodes Sequence through embedding matrix $A$ and mapping the Node Position Sequence via matrix $B$. Specifically, we prepend a Class Token Embedding and its corresponding Node Position Embedding to both embedding sequences to establish a fixed prediction anchor. 

The final embedding $E_g$ for token $g$ integrates these components additively: $E_g = A_g + B_g$. Crucially, we prepend the sequence with the navigation signal embedding $E_{\text{sig}} = h_Q$ (extracted by the QSExtractor) to provide target structural signals for subsequent decoder layers. 
The decoder input embedding forms the sequence $X_{\text{in}} = [E_{[\text{Sig}]}, E_{[\text{Cls}]}, E_0, \ldots, E_{l-1}] \in \mathbb{R}^{(l+2) \times d}$, where $l$ is the length of the subgraph’s (semi-)Eulerian node sequence, with BERT-style positional embeddings added.

\textbf{Masked Nodes Decoder.}
\label{subsec:Decoder}
The decoder employs a bidirectional Transformer Decoder architecture to propagate structural constraints derived from the Euler-guided sequence. Given the embedded input $X_{in} \in \mathbb{R}^{(l+2) \times d}$, the $K$ stacked layers iteratively transform the hidden representations in a hierarchical manner. For the $k$-th layer, with $H^{(0)} = X_{in}$ and $H^{(k-1)}$ as input, the forward pass is defined as:
\begin{equation}
\begin{aligned}
H_{\text{att}}^{(k)} &= \text{LayerNorm}\left(\text{MultiHead}(H^{(k-1)}) + H^{(k-1)}\right), \\
H^{(k)} &= \text{LayerNorm}\left(\text{FFN}(H_{\text{att}}^{(k)}) + H_{\text{att}}^{(k)}\right),
\end{aligned}
\end{equation}
where $\text{MultiHead}(\cdot)$ denotes the multi-head self-attention and $\text{FFN}(\cdot)$ is a position-wise feed-forward network. 
 
\textbf{Prediction.}
The final prediction is derived from the anchor class token's hidden representation after passing through the decoder layers. Let ${h}_{[Cls]} \in \mathbb{R}^d$ denote the output embedding of the class token. 
We project it to the vocabulary space via a linear transformation followed by softmax:
\begin{equation}  
\label{eq.prediction} 
P = \text{Softmax}\left(W {h}_{[Cls]} + b\right)  ,  
\end{equation}  
where $W \in \mathbb{R}^{\lvert V_n \rvert \times d}$ and $b \in \mathbb{R}^{\lvert V_n \rvert}$ are learnable parameters. The output $P \in \mathbb{R}^{\lvert V_n \rvert} $ forms a probability distribution over all candidate nodes. 

\subsection{Training Strategy}
\textbf{Data Preprocessing.} 
We adopt Masked Node Generation (MNG) for self-supervised training. In each epoch, for every node $v$ in the data graph, we sample a variable-size query subgraph  centered at $v$ and convert it into a (semi-)Eulerian path sequence. During training, we randomly mask multiple nodes in the sequence and select one masked node as the prediction target. Details can be found in Appendix B.4.\\
\textbf{Loss Function.}
The model is trained using cross-entropy loss between the predicted distribution $P$ and the ground-truth node label. For a target node with index $t$ in the candidate node set $V_n$, the loss is:
\begin{equation}  
\label{eq.loss}
\mathcal{L}_{\text{MNG}} = -\log P_t,
\end{equation}
The training objective is to maximize the probability assigned to the true match, thereby guiding the navigator to prioritize correct candidates.

\begin{figure}[t]
    \centering 
    \includegraphics[width=0.95\columnwidth]{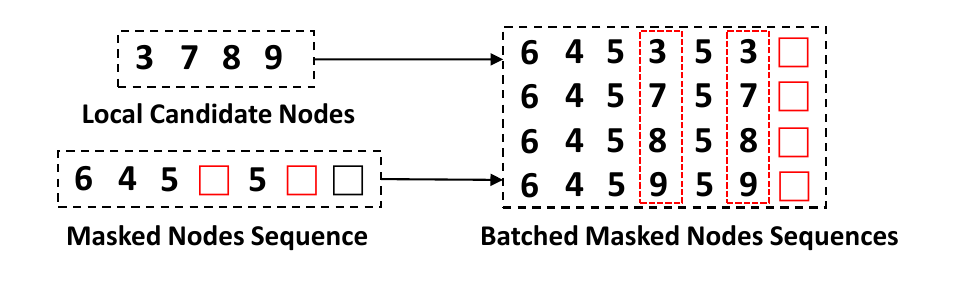}
    \caption{Local Candidate Nodes Batching Strategy.}
    \label{fig:batch}
\end{figure}

\subsection{Plug-and-play Deployment of NeuGN}
This section presents how NeuGN integrates with other algorithms to endow their enumeration phases with intelligent search capabilities.\\
\textbf{Query Graph Preprocessing.} As illustrated in Algorithm~\ref{alg:subgraph}, QSExtractor preprocesses query graphs to extract their navigation signals before the filtering phase. For streaming query inputs, we employ batch-based parallel processing to utilize the powerful GPUs.\\  
\textbf{Score Calculation and Candidate Ranking.} GGNavigator operates on the local candidate nodes $\mathcal{C}$ generated by the enumerator. A prioritized ranking score is formulated to rank these candidates for effective matching. Given a candidate $c \in \mathcal{C}$, its confidence score is defined as:  
\begin{equation}
\label{eq.score_function} 
\text{Conf}(c) = \sum_{u \in \mathcal{C}} \mathbb{I}(P(c) > P(u)),  
\end{equation}  
where $\mathbb{I}(\cdot)$ is the indicator function. The local candidate set is then reordered in descending confidence order:  
\begin{equation}  
\label{eq.sort}  
\text{SortedCandidates} = \text{argsort}(\{\text{Conf}(c) | c \in \mathcal{C}\}).  
\end{equation}  
Crucially, NeuGN only reorders candidates in $\mathcal{C}$ , without additional pruning. This preserves the \textit{completeness guarantee} of the base algorithm (Proof: Appendix A).\\
\textbf{Local Candidates Batching Strategy.}
To mitigate the computational cost of deep network inference, we introduce a batching strategy leveraging GPU parallelism. During enumeration, the local candidate nodes for the next matching step are embedded into the Masked Nodes Sequence to form a batch of input sequences. These sequences share a common partial substructure but differ in the local nodes to be matched. Formally, the batched input is defined as:
\begin{equation}
\label{eq.batch}
\mathcal{B} = \left\{ \text{Construct}(\mathcal{M}_{\text{partial}} \cup \{c\}) \, \vert \, c \in \mathcal{C}_{\text{next}} \right\},
\end{equation}  
where $\mathcal{M}_{\text{partial}}$ denotes the current partial match and $\mathcal{C}_{\text{next}}$ is the local candidates set. The construction method is detailed in Figure~\ref{fig:batch}. The model performs parallel inference on $\mathcal{B}$, outputting confidence scores $\{\text{Conf}(c) \mid c\in\mathcal{C}_{\text{next}}\}$ for all masked nodes sequences in a single pass. Subsequently, we maintain a priority list for $\{\text{Conf}(c) \mid c\in\mathcal{C}_{\text{next}}\}$ to facilitate the intricate process of enumeration in subsequent steps.

\begin{table}[t]
    \small
    \centering
    \begin{tabular}{ccccr} 
        \toprule 
        \textbf{Dataset} & $|V|$ & $|E|$ & $|L|$ & \textbf{$\bar{d}$}\\ 
        \midrule 
        Hamster & 2,421 & 16,621 & 16 & 13.73 \\
        LastFM & 7,624 & 27,806 & 18 & 7.29 \\
        WikiCS & 11,701 & 215,603 & 10 & 36.85 \\
        NELL & 65,755 & 125,775 & 105 & 3.83 \\
        DBLP & 317,080 & 1,049,866 & 15 & 6.62 \\
        YouTube & 1,134,890 & 2,987,624 & 25 & 5.27 \\
        \midrule
    \end{tabular}
    \caption{Dataset statistics.}
    \label{tab:datasets}
\end{table}

\begingroup
\begin{table*}[t]
\small
\centering
\begin{tabular}{crrrrrrrrrrrr}
\toprule
 & \multicolumn{2}{c}{Hamster} & \multicolumn{2}{c}{LastFM} & \multicolumn{2}{c}{Wikics} & \multicolumn{2}{c}{Nell} & \multicolumn{2}{c}{DBLP} & \multicolumn{2}{c}{YouTube} \\ \cmidrule(lr){2-3}\cmidrule(lr){4-5} \cmidrule(lr){6-7} \cmidrule(lr){8-9} \cmidrule(lr){10-11} \cmidrule(lr){12-13} 
 & \multicolumn{1}{l}{Dense} & \multicolumn{1}{l}{Sparse} & \multicolumn{1}{l}{Dense} & \multicolumn{1}{l}{Sparse} & \multicolumn{1}{l}{Dense} & \multicolumn{1}{l}{Sparse} & \multicolumn{1}{l}{Dense} & \multicolumn{1}{l}{Sparse} & \multicolumn{1}{l}{Dense} & \multicolumn{1}{l}{Sparse} & \multicolumn{1}{l}{Dense} & \multicolumn{1}{l}{Sparse} \\ \midrule
QSI & 4974 & 218 & 3154 & 668 & 37562 & 2168 & 1636 & 452 & 7750 & 2334 & 47198 & 564 \\
+NeuGN & 947 & 120 & 132 & 110 & 74 & 58 & 80 & 67 & 1448 & 1006 & 9433 & 235 \\
Improv. & 81.0\% & 45.0\% & 95.8\% & 83.5\% & 99.8\% & 97.3\% & 95.1\% & 85.2\% & 81.3\% & 56.9\% & 80.0\% & 58.3\% \\ \hline
GQL & 165 & 49 & 136 & 32 & 1566 & 108 & 69 & 25 & 55 & 61 & 563 & 32 \\
+NeuGN & 39 & 32 & 22 & 21 & 21 & 25 & 20 & 20 & 21 & 37 & 183 & 21 \\
Improv. & 76.4\% & 34.7\% & 83.8\% & 34.4\% & 98.7\% & 76.9\% & 71.0\% & 20.0\% & 61.8\% & 39.3\% & 67.5\% & 34.4\% \\ \hline
CFL & 1266 & 37 & 4666 & 34 & 18482 & 60 & 14332 & 33 & 51 & 60 & 10978 & 26 \\
+NeuGN & 236 & 30 & 76 & 24 & 44 & 22 & 2348 & 25 & 22 & 46 & 2433 & 23 \\
Improv. & 81.4\% & 18.9\% & 98.4\% & 29.4\% & 99.8\% & 63.3\% & 83.6\% & 24.2\% & 56.9\% & 23.3\% & 77.8\% & 11.5\% \\ \hline
VF3 & 402 & 36 & 551 & 51 & 2235 & 81 & 254 & 30 & 468 & 144 & 728 & 32 \\
+NeuGN & 61 & 29 & 28 & 38 & 53 & 34 & 23 & 24 & 84 & 86 & 271 & 22 \\
Improv. & 84.8\% & 19.4\% & 94.9\% & 25.5\% & 97.6\% & 58.0\% & 90.9\% & 20.0\% & 82.1\% & 40.3\% & 62.8\% & 31.3\% \\ \hline
CECI & 162 & 36 & 163 & 31 & 675 & 36 & 3712 & 32 & 106 & 70 & 916 & 65 \\
+NeuGN & 61 & 30 & 22 & 22 & 35 & 24 & 35 & 25 & 29 & 36 & 235 & 38 \\
Improv. & 62.3\% & 16.7\% & 86.5\% & 29.0\% & 94.8\% & 33.3\% & 99.1\% & 21.9\% & 72.6\% & 48.6\% & 74.3\% & 41.5\% \\ \hline
CaLiG & 276 & 42 & 289 & 43 & 1136 & 69 & 8105 & 65 & 98 & 85 & 1045 & 57 \\
+NeuGN & 89 & 32 & 77 & 40 & 51 & 35 & 1214 & 35 & 34 & 43 & 405 & 33 \\
Improv. & 67.8\% & 23.8\% & 73.4\% & 7.0\% & 95.5\% & 49.3\% & 85.0\% & 46.2\% & 65.3\% & 49.4\% & 61.2\% & 42.1\% \\ \hline
RLQVO & 2074 & 344 & 1486 & 206 & 17082 & 1262 & 1446 & 44 & 2282 & 1560 & 18018 & 616 \\
+NeuGN & 540 & 124 & 74 & 88 & 70 & 58 & 94 & 27 & 504 & 336 & 3879 & 335 \\
Improv. & 74.0\% & 64.0\% & 95.0\% & 57.3\% & 99.6\% & 95.4\% & 93.5\% & 38.6\% & 77.9\% & 78.5\% & 78.5\% & 45.6\% \\ \hline
RSM & 2358 & 189 & 952 & 173 & 12678 & 1551 & 2556 & 42 & 1596 & 1326 & 9967 & 456 \\
+NeuGN & 489 & 78 & 53 & 42 & 59 & 67 & 267 & 25 & 173 & 345 & 2940 & 189 \\
Improv. & 79.3\% & 58.7\% & 94.4\% & 75.7\% & 99.5\% & 95.7\% & 89.6\% & 40.5\% & 89.2\% & 74.0\% & 70.5\% & 58.6\% \\ \midrule
Average. & \textbf{75.9\%} & \textbf{35.1\%} & \textbf{90.3\%} & \textbf{42.7\%} & \textbf{98.2\%} & \textbf{71.2\%} & \textbf{88.5\%} & \textbf{37.1\%} & \textbf{73.4\%} & \textbf{51.3\%} & \textbf{72.4\%} & \textbf{40.4\%} \\ \bottomrule
\end{tabular}
\caption{Performance Comparison on FMS (lower is better) Across NeuGN-Enhanced Subgraph Matching Algorithms. }
\label{tab:matchn}
\end{table*}
\endgroup

\section{Experiments}
In this section, we evaluate the proposed NeuGN by addressing the following research questions:\\
\textbf{RQ1:} How does NeuGN reduce First Match Steps (FMS) across diverse graph datasets?\\  
\textbf{RQ2:} How do individual components of NeuGN contribute to its performance efficacy?\\
\textbf{RQ3:} How does the quality of neural guidance scale with NeuGN's navigation depth in the enumeration tree?\\
\textbf{RQ4:} How does the performance of NeuGN vary with increasing query graph size?\\
\textbf{RQ5:}  Does neural navigation effectively optimize early-stage enumeration efficiency in subgraph matching?\\

\subsection{Experimental setup}
\textbf{Datasets.} We evaluate our method on six real-world benchmark datasets: Hamster~\cite{kunegis2013konect}, LastFM~\cite{lastfm}, WikiCS~\cite{mernyei2020wikics}, NELL~\cite{carlson2010nell1}, DBLP~\cite{yang2012defining}, and YouTube~\cite{yang2012youtube}. Statistics are summarized in Table~\ref{tab:datasets}. More details of the datasets can be found in Appendix C.1.\\
\textbf{Baselines.} In order to show the effectiveness of NeuGN, we conduct experiments against 8 advanced baselines which include both traditional and learning-enhanced methods: QSI~\cite{shang2008qsi}, GQL~\cite{he2008gql},  CFL~\cite{bi2016cfl}, VF3~\cite{carletti2017vf3}, CECI~\cite{bhattarai2019ceci},  CaLiG~\cite{yang2023calig}, RLQVO~\cite{wang2022RLQVO} and RSM~\cite{li2025rsm}. \\
\textbf{Metrics.} (1) \textbf{First Match Steps (FMS)}: number of enumeration steps to find the first match; lower FMS indicates better navigation. (2) \textbf{Matches Per Second (MPS)}: number of matches returned per second, also known as Embeddings Per Second (EPS)~\cite{zhang2024survey}; higher MPS indicates better throughput.\\
\textbf{Implementation.}
We implement the offline training module in Python using PyTorch, while the online query evaluation is implemented in C++ with LibTorch for low-latency inference. 
Unless otherwise specified, we use the following default settings: query graphs with 20 nodes, a query stream of 200 graphs, and a navigation depth fixed at 10. Further implementation details and experiments on training and inference efficiency are provided in Appendix~C.2 and C.3.

\subsection{Performance Comparison}
\subsubsection{RQ1: FMS Comparison Across Algorithms and Datasets.} 
\begin{figure}[h]
    \centering 
    \includegraphics[width=0.98\columnwidth]{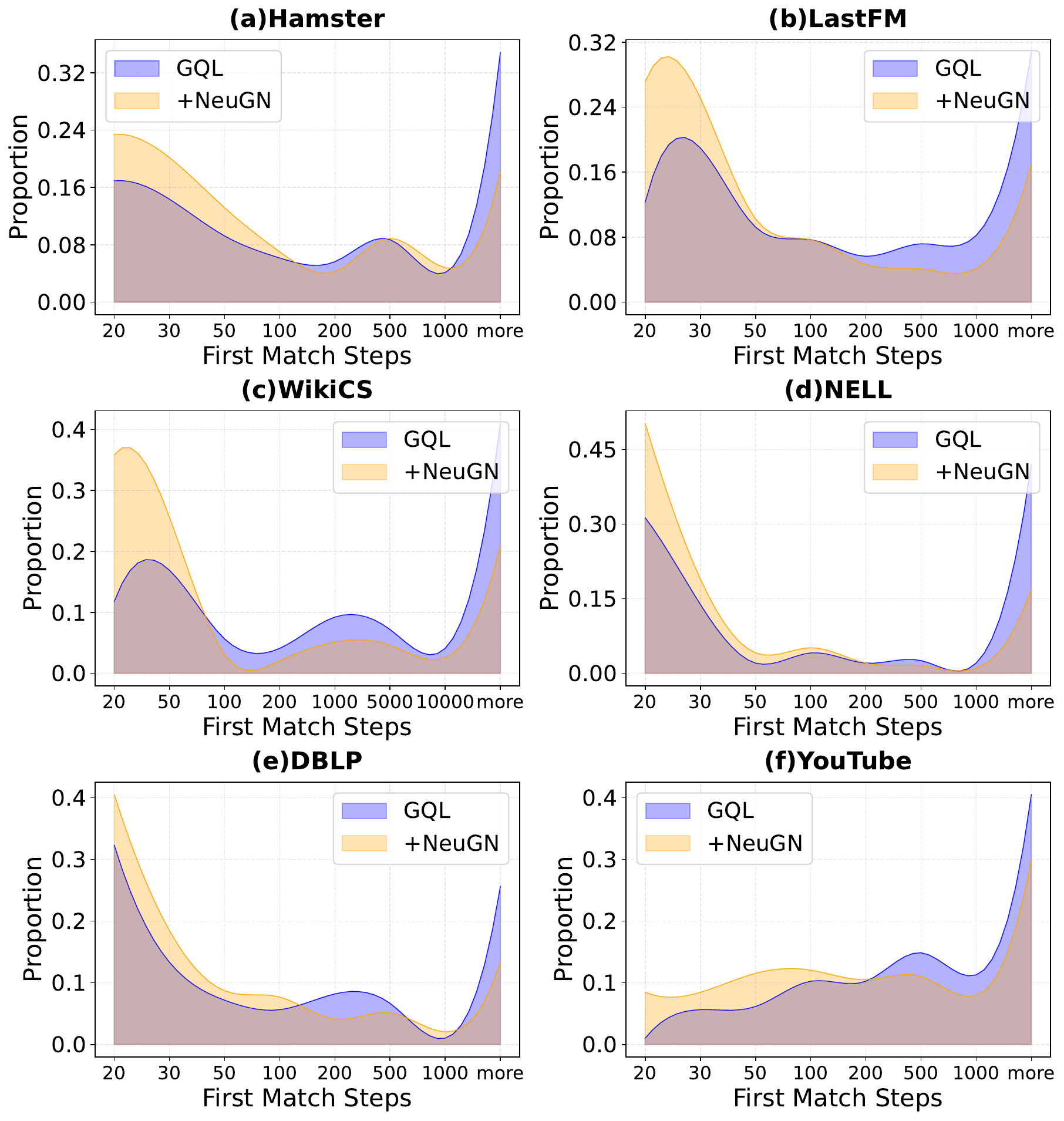}
    \caption{Distribution of FMS Across All Queries.}
    \label{fig:flow}
\end{figure}

We quantitatively evaluated the navigational intelligence of NeuGN by analyzing the FMS. Subgraph density was categorized as \textbf{sparse} (avg. degree $d_{avg} < 3$) or \textbf{dense} ($d_{avg} \geq 3$) to assess the performance under varying topological complexity. Crucially, we report the median (rather than mean) to avoid distortion from extreme outliers inherent to backtracking-based searches. As shown in Table~\ref{tab:matchn}, NeuGN achieves significant FMS reductions across all datasets for both density categories.
The reduction magnitude is substantially larger for dense subgraphs, where traditional heuristics methods suffer from exponentially growing branches, and the learned prioritization of NeuGN correctly focuses on promising nodes early. In addition to Table~\ref{tab:matchn}, Figure~\ref{fig:flow} visualizes the distribution of FMS across all queries between GQL and NeuGN-enhanced GQL. The leftward shift in curves of NeuGN shows its consistent concentration of matches at lower step counts, confirming the ability of the NeuGN framework to inject structural awareness into enumeration optimization.

\subsubsection{RQ2: Ablation Study.} 
To evaluate the contribution of individual components in NeuGN, we design three ablated variants: (1) disabling the QSExtractor (w/o Extr.), (2) replacing the GGNavigator with a multi-layer perceptron (MLP Navi.), and (3) substituting the (semi-)Eulerian node path in the GGNavigator with a random walk path (RW Navi.). All experiments are conducted within the CECI framework, with FMS as the evaluation metric. Disabling the QSExtractor leads to a significant increase in the FMS across all datasets (Table~\ref{tab:ablation}), as the target navigation signal from the query graph is no longer available, confirming its critical role in navigation quality. Replacement of the navigator with an MLP also degrades performance. Unlike our generative navigator, which jointly encodes structural features and evolving partial match states, the MLP operates on static input and lacks adaptation to current matching progress. The variant using a random walk path performs slightly better than the MLP-based version by capturing dynamics of partial matches, but it underperforms compared to the full NeuGN model. Unlike (semi-)Eulerian paths, random walks do not provide a lossless representation of the subgraph structure. This impairs the model’s ability to perceive structure and leads to less effective navigation.

\begin{table}[]
\small
\centering
\setlength{\tabcolsep}{1.1mm}
\begin{tabular}{l crlrlrl}
\toprule
 \textbf{Datasets}& \textbf{NeuGN} & \multicolumn{2}{c}{\textbf{w/o Extr.}} & \multicolumn{2}{c}{\textbf{MLP Navi.}} & \multicolumn{2}{c}{\textbf{RW Navi.}} \\
\hline
Hamster  & \textbf{61}  & 170 &↑179\%  & 91 &↑49\%  & 74 &↑21\% \\
LastFM   & \textbf{22}  & 226 &↑927\%  & 45 &↑105\% & 32 &↑45\% \\
WikiCS   & \textbf{35}  & 378 &↑980\%  & 66 &↑89\% & 48 &↑37\% \\
NELL     & \textbf{35}  & 2,699 &↑7,611\%& 209 &↑497\%& 86 &↑146\% \\
DBLP     & \textbf{29}  & 91 &↑213\%   & 44 &↑52\%  & 32 &↑10\% \\
YouTube  & \textbf{235} & 767 &↑226\%  & 422 &↑80\% & 316 &↑34\% \\
\hline
\end{tabular}
\caption{Ablation Study of NeuGN Components on FMS ($\uparrow$ indicates performance degradation).}
\label{tab:ablation}
\end{table}

\subsubsection{RQ3: Impact of Navigation Depth.}
\begin{figure}[h]
    \centering 
    \includegraphics[width=0.97\columnwidth]{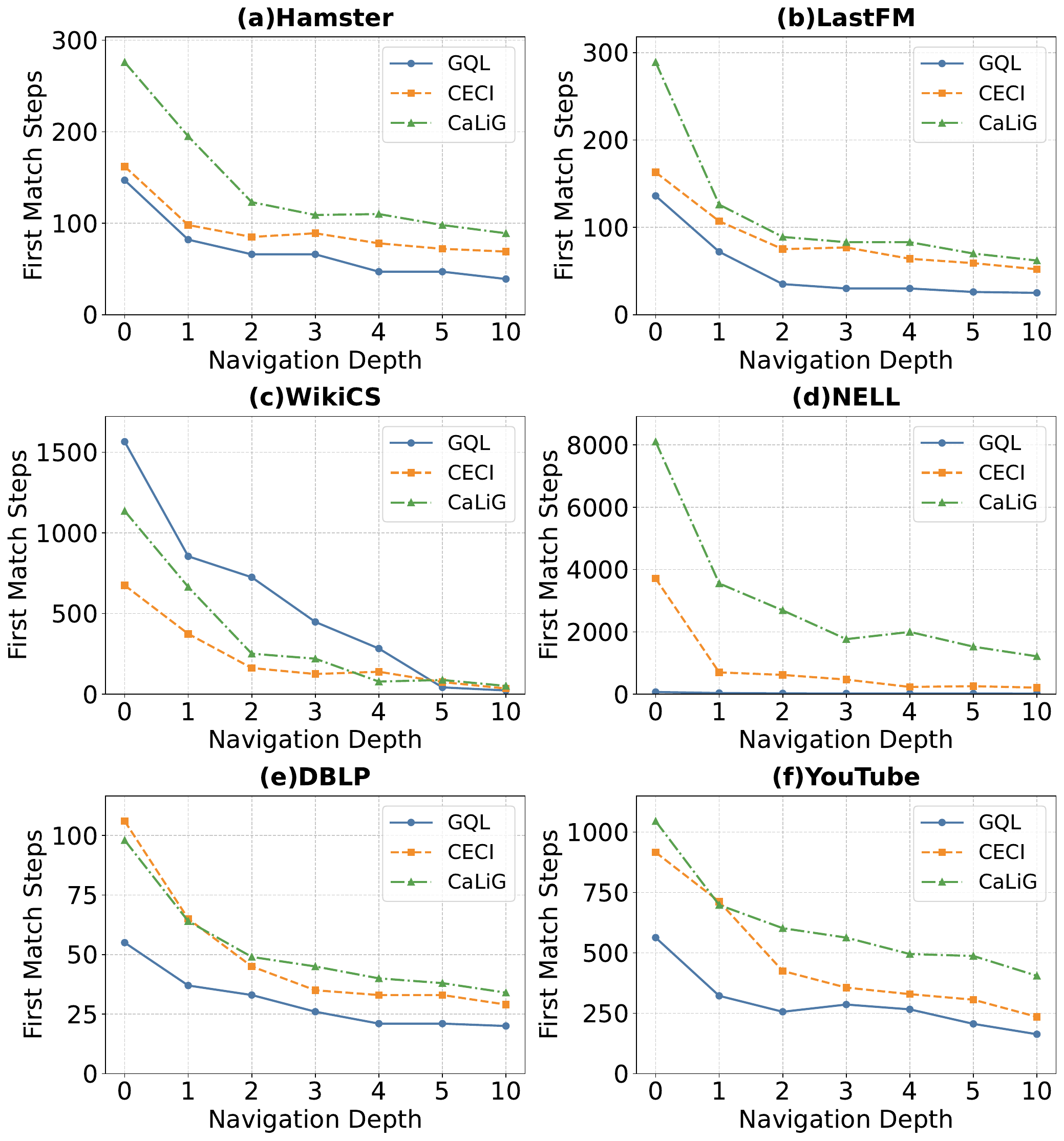}
    \caption{Impact of Navigation Depth on FMS.}
    \label{fig:depth}
\end{figure}

To evaluate the navigation mechanism of NeuGN, we study its effectiveness when applied to different depths in the enumeration tree. Figure~\ref{fig:depth} shows the impact of navigation depth. The $x$-axis denotes the maximum depth to which NeuGN provides navigation, ranging from 0(no navigation) to 10; beyond this depth, the baseline ordering is used. The $y$-axis reports the median FMS across queries. For all datasets and algorithms, FMS decreases significantly as the navigation depth increases. The most substantial improvements occur in the early stages (depths from 0 to 3), indicating NeuGN quickly steers the search toward high-probability matching nodes.

\begin{table}[]
\small
\centering
\begin{tabular}{ccrrrr}
\toprule
\multicolumn{1}{l}{}    &         & 8 N & 16 N & 24 N & 32 N \\ \hline
\multirow{3}{*}{WikiCS} & CECI    & 19      & 108       & 2,707      & 254,460     \\
                        & +NeuGN  & 8       & 12       & 101       & 8,079       \\
                        & improv. & \textbf{57.9\%}  & \textbf{85.2\%}   &\textbf{96.3\%}   & \textbf{96.8\%}   \\ \hline
\multirow{3}{*}{NELL}   & CECI    & 9       & 328       & 10,092      & 103,473      \\
                        & +NeuGN  & 8       & 33       & 992       & 8,715       \\
                        & improv. & \textbf{11.1\%}  & \textbf{89.9\%}   & \textbf{90.2\%}   & \textbf{91.6\%}   \\ \hline
\end{tabular}
\caption{FMS of NeuGN-Enhanced CECI Across Different Query Sizes.}
\label{tab:nodes}
\end{table}

\subsubsection{RQ4: Generalization to Different Query Sizes.}
To evaluate the scalability of NeuGN, we conduct experiments on query graphs with 8, 16, 24, and 32 nodes. As shown in Table~\ref{tab:nodes}, the NeuGN-enhanced CECI framework consistently outperforms the baseline, with performance improvements becoming more pronounced as query size increases. This trend demonstrates that NeuGN effectively prioritizes high-probability candidates, particularly in larger and more complex search spaces arising from larger query graphs.

\begin{table}[]
\small
\centering
\begin{tabular}{lcccc}
\toprule
 & \multicolumn{1}{c}{Hamster} & \multicolumn{1}{c}{LastFM} & \multicolumn{1}{c}{WikiCS} & \multicolumn{1}{c}{NELL} \\ \hline
GQL & 4.29E+05 & 8.22E+05 & 3.06E+05 & 4.58E+05 \\
+NeuGN & 5.02E+05 & 1.14E+06 & 4.67E+05 & 5.98E+05 \\
Improv. & \textbf{17.0\%} & \textbf{39.0\%} & \textbf{52.4\%} & \textbf{30.5\%} \\ \hline
CECI & 5.15E+06 & 5.04E+06 & 4.93E+06 & 4.79E+06 \\
+NeuGN & 5.62E+06 & 5.94E+06 & 6.79E+06 & 5.76E+06 \\
Improv. & \textbf{9.2\%} & \textbf{17.8\%} & \textbf{37.7\%} & \textbf{20.2\%} \\ \hline
\end{tabular}
\caption{MPS in Time-Bounded Subgraph Enumeration.}
\label{tab:mps}
\end{table}

\subsubsection{RQ5: Acceleration of Early Convergence.}
To evaluate whether NeuGN accelerates early convergence by prioritizing high-probability candidate nodes, we impose a 1-second time budget for enumeration on query graphs of size 32, with navigation depth fixed at 16. Matching throughput, measured by MPS, is significantly improved under NeuGN. As shown in Table~\ref{tab:mps}, NeuGN achieves substantial efficiency gains during the initial enumeration phase. The performance improvement from structural navigation outweighs the inference latency, resulting in a positive impact on time-bounded subgraph enumeration. More details and additional experiment results are provided in Appendix~C.3.

\section{Conclusion}
In this paper, we propose NeuGN, a novel plug-and-play neural navigation framework for subgraph matching that introduces a generative method to navigate the enumeration phase.
To the best of our knowledge, we present the first approach to explicitly guide subgraph enumeration through neural-based navigation in exact subgraph matching. Comprehensive empirical evaluations demonstrate that NeuGN achieves a reduction of up to 98.2\% in FMS and substantially accelerates early convergence compared to state-of-the-art methods across six real-world datasets. This significant efficiency gain paves the way for future research into more advanced neural navigation mechanisms and their application to increasingly complex subgraph matching tasks. 

\section{Acknowledgments}
This work is supported in part by the Starry Night Science Fund of Zhejiang University Shanghai Institute for Advanced Study (Grant No. SN-ZJU-SIAS-001), Zhejiang Provincial Natural Science Foundation of China (Grant No. LMS25F020012), the Hangzhou Joint Fund of the Zhejiang Provincial Natural Science Foundation of China under Grant No.LHZSD24F020001, Zhejiang Province High-Level Talents Special Support Program "Leading Talent of Technological Innovation of Ten-Thousands Talents Program" (No.2022R52046), the Fundamental Research Funds for the Central Universities (No.226-2024-00058), and the advanced computing resources provided by the Supercomputing Center of Hangzhou City University.

\bibliography{aaai2026}

\appendix
\renewcommand{\thesection}{\Alph{section}}
\renewcommand{\thesubsection}{\thesection.\arabic{subsection}}
\setcounter{secnumdepth}{2}

\section*{Technical Appendix}

In this appendix, we provide a comprehensive elaboration of the methodologies, experimental details, and additional insights that support the findings presented in the main manuscript. The appendix is structured into the theoretical analysis of NeuGN (\textit{Section A}), details of the proposed method (\textit{Section B}), details of the experiments (\textit{Section C}), and further discussions on limitation and future work (\textit{Section D}). Furthermore, our source code is provided in the Supplementary Materials.

\section{Theoretical Analysis of NeuGN}

This appendix provides the theoretical underpinning of NeuGN by formally proving its \emph{completeness guarantee}, with a proof sketch presented below. Section~A.1 demonstrates that the NeuGN-enhanced enumerator outputs exactly the same match set as any sound‐and‐complete baseline searcher (e.g., QSI, GQL, VF3, CECI, etc.) whose pruning routine obeys the standard safe–pruning requirement: no candidate that could still be extended to a full match is discarded before it is generated and tested, irrespective of the order in which its siblings are explored.  The argument rests on two observations: (i) NeuGN is purely permutational, it reorders but never removes elements from the candidate list at each depth, so candidate coverage is preserved; (ii) classical order-robust pruning rules satisfy the safe–pruning property. Combining these facts, we prove that every root‐to‐leaf path yielding a valid match in the baseline search tree remains reachable when NeuGN dictates the visitation order, guaranteeing that no admissible match is lost even though the traversal order changes.

\subsection{Completeness Guarantee}

We prove that inserting NeuGN, which only reorders local candidate nodes, preserves the soundness and completeness of the underlying back-tracking enumerator.

\subsubsection{Setup and Assumptions.}

Let $Q=(V_Q,E_Q)$ be the query graph, $G=(V_G,E_G)$ the data
graph, and $\phi=(u_1,\dots,u_{|V_Q|})$ a fixed visiting order of
query vertices.
At depth $i$ the current partial injective mapping is
$
  M_i=\{(u_j,v_j)\mid j\le i\},
$
and the next candidate multiset is
$\mathcal{C}_{M_i}(u_{i+1})$.
Depth-first search (DFS) expands candidates one by one; a
full-depth leaf is a valid match.

\paragraph{Deterministic baseline order.}
The baseline enumerator explores every
$\mathcal{C}_{M_i}$ in a fixed deterministic order
$\pi_0$ (e.g.\ ascending vertex id).

\paragraph{Permutation policy.}
NeuGN replaces $\pi_0$ by a learned permutation
$\pi_N(M_i)$ while keeping

\begin{enumerate}[label=(\alph*),leftmargin=14pt]
  \item the same visiting order~$\phi$ of
query vertices and candidate generation,
  \item the same pruning predicate \texttt{prune}.
\end{enumerate}

\medskip\noindent
We adopt two explicit assumptions that hold for most classical
algorithms (e.g.\ QSI, GQL, VF3, CECI).

\begin{description}[leftmargin=14pt]
  \item[(A1) Order-robust safe pruning.]
If a candidate $(u_{i+1},v)$ can still be extended to a full
        match, the pruning predicate \texttt{prune} never deletes
        the subtree rooted at that child \emph{before} the child has
        been generated and tested, no matter in which order the other
        siblings of $u_{i+1}$ are explored.

  \item[(A2) Admissible permutation.]
    For every state $M_i$, $\pi(M_i)$ is a complete
    bijection on~$\mathcal{C}_{M_i}(u_{i+1})$; i.e.\
    all candidates appear exactly once.
    (Sampling top‐$k$ or dropping low-score elements would violate A2.)
\end{description}

Under assumptions (A1) and (A2), the search trees associated with both the baseline and NeuGN-enhanced enumeration procedures are finite and acyclic, thereby guaranteeing the termination of the depth-first search process.

\subsubsection{Lemma 1 (Solution-Path Preservation)}
\label{lem:solpath}
For any admissible permutation policy $\pi$, \emph{every} the root-to-leaf path that produces a valid match under the baseline order $\pi_{0}$ is also generated under $\pi$.

\begin{proof}
Proceed by induction along a fixed baseline solution path
$M_{0}\!\to\!M_{1}\!\to\dots\!\to\!M_{|V_Q|}$.
Assume the common prefix $M_{0},\dots,M_{i}$ has already been produced
by the NeuGN order $\pi$. The next mapping on the path is
$M_{i+1}=M_{i}\cup\{(u_{i+1},v_{i+1})\}$.
Because of (A2), $v_{i+1}$ still appears in the permuted candidate
list, and by (A1) the subtree below that child is not pruned
before the candidate is tested.  Hence DFS will visit $M_{i+1}$ and
the induction continues.

\end{proof}

\subsubsection{Theorem 1 (Soundness and Completeness)}
\label{thm:complete}

If the baseline enumerator (with $\pi_{0}$) is sound and complete,
then, under assumptions (A1) and (A2), the
NeuGN-enhanced enumerator (with $\pi_{N}$) is also sound and
complete.

\begin{proof}
\emph{Soundness.} Soundness is unchanged because all feasibility checks and the
pruning predicate themselves are left intact.
\emph{Completeness.}
Let $M^{\star}$ be any match output by the baseline, i.e.\ a leaf
of its search tree.  By Lemma~\ref{lem:solpath}, the same solution path
exists in the NeuGN search tree.  Because \textsc{prune} is
safe according to (A1), no ancestor of $M^{\star}$ is removed
before it is reached.  Finite DFS therefore eventually visits
$M^{\star}$, so the match is output as well.
\end{proof}

\paragraph{Corollary 1.}
Any exact subgraph matcher whose pruning routine satisfies the
order-robust safe pruning property (A1) can be combined with NeuGN without loss of completeness, irrespective of auxiliary
data structures (e.g.\ failing sets, hash tables) that may depend on
enumeration order.

\smallskip
This completes the proof.

\section{Details of the NeuGN Method}

\begin{algorithm}[t]
\small
\caption{NeuGN Framework for Subgraph Matching.} 
\label{alg:subgraph}
\KwIn{A query graph \( Q \) and a data graph \( G \).} 
\KwOut{All the matches of \( Q \) into \( G \).}    

\BlankLine 
\( \bm{h_Q} \leftarrow \textbf{QSExtractor}\bm{(Q)} \;\)  

\( C \leftarrow \text{FilterNodes}(Q, G)\)\; 

\( \varphi \leftarrow \text{GenerateOrder}(Q, G, C)\)\;

\(\text{Enumerate}(Q, G, C, \varphi, \{ \}, \bm{h_Q}, 0)\)\; 

\BlankLine
\SetKwProg{Fn}{Function}{:}{} 
\Fn{\text{Enumerate}(Q, G, C, \(\varphi\), M, \(\bm{h_Q}\), i)}{  
    \If{Termination condition met}{ 
        \text{output} \text{TerminateFunc}\((Q, G, C, \varphi, M)\)\;
    }
    \( u \leftarrow \text{SelectNextQueryNode}(Q, G, C, \varphi, M) \)\;
    
    \( C_M(u) \leftarrow \text{ComputeLocalCandidates}(Q, G, C, \varphi, M, i) \)\;
    
    \( \bm{Conf} \leftarrow \textbf{GGNavigator}\bm{(Q}, \bm{G}, \bm{C}, \bm{h_Q}, \bm{\varphi}, \bm{M)} \)  
    
    \( \bm{C_M(u)} \leftarrow \textbf{Sort}\bm{(C_M(u)}, \bm{Conf)} \)\;  
    
    \ForEach{\( v \in C_M(u) \)}{

        \(\text{PruningFunc}(Q, G, C, \varphi, M)\)\;
    
        Extend \( M \) by \( (u, v) \)\;
        
        \(\text{Enumerate}(Q, G, C, \varphi, M, i + 1)\)\; 
        
        Delete \( (u, v) \) from \( M \)\;
        
    }
}
\end{algorithm}

\subsection{Overall Algorithm Description} 
The general framework of filtering-ordering-enumeration enhanced by NeuGN is outlined in Algorithm~\ref{alg:subgraph} in the main text, with a detailed description provided in this section. 

\textbf{Extracting Query Structure.} The algorithm first extracts a structural signal $h_Q$ of the query graph $Q$ using the proposed \textbf{QSExtractor}  (line 1). This signal serves as a compact representation of $Q$’s structure patterns, enabling the GGNavigator to effectively align the evolving partial matches in the data graph $G$ with the target substructure.

\textbf{Filtering nodes.} The algorithm then generates candidate sets by filtering nodes (line 2). Compared to traversing the entire graph, the filtering phase identifies potential matches for each query node and eliminates data graph nodes that cannot satisfy the matching constraints. The overall candidate set is denoted by $C$, and the candidates for a specific query node $u$ are denoted by $C(u)$.

\textbf{Matching Order.} The algorithm then generates a matching order (line 3). A matching order $\varphi$ of $Q$ refers to a permutation of $V_Q$ , determining the order in which nodes are explored during the search process. The index of vertex $u$ in $\varphi$ is denoted by $\varphi(u)$.

\textbf{Enumeration Process.} The algorithm finally conducts enumeration (line 5). The core enumeration process first computes the local candidate set $C_M(u)$ for the next query node (line 9). After obtaining $C_M(u)$, NeuGN employs the \textbf{GGNavigator} to score and rerank the candidates (lines 10–11). followed by extending partial matches  (line 14), and recursively carries out the computations (line 15). The basic termination condition (line 6) is met when $i = |V_Q|$, indicating a complete match has been found. Pruning operations, which can eliminate invalid partial matches early, are applied during the enumeration (lines 13, 16). Some methods further optimize the process by using additional indices to avoid unnecessary backtracking.

\subsection{(Semi-)Eulerian Node Path} 

To enable lossless and reversible graph serialization via (semi-)Eulerian paths, we employ the \textit{eulerize} algorithm from NetworkX~\cite{hagberg2008networkx}, which transforms an arbitrary connected undirected graph into an Eulerian graph by duplicating edges. The algorithm first identifies all vertices with odd degree. If none exist, the graph is already Eulerian. Otherwise, it computes the shortest paths between all pairs of odd-degree vertices and constructs an auxiliary complete graph $G_p$. A minimum-weight perfect matching in $G_p$ identifies the optimal set of paths whose edges are duplicated, ensuring all vertices attain even degree with minimal edge addition. The Eulerization process has a polynomial time complexity of $O(|V|^3)$, which is dominated by the minimum-weight perfect matching computation on the set of odd-degree vertices.

Crucially, this Eulerization step is performed only on the query graph $Q$, which is typically small and fixed in size (e.g., $|V_Q| \ll |V_G|$). In contrast, the dominant cost of subgraph matching arises from the search over the data graph $G$, which has worst-case time complexity $O(|V_G|^{|V_Q|})$. Given that the Eulerization cost is polynomial in $|V_Q|$ and independent of $|V_G|$, it constitutes a negligible preprocessing overhead and is therefore fully acceptable within the overall computational framework.

\begin{figure}[ht]
  \centering
  \includegraphics[width=0.5\textwidth]{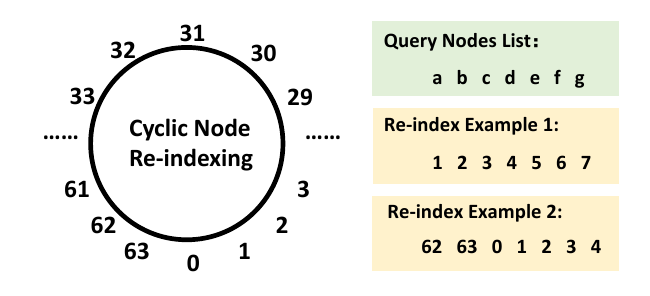}
  \caption{Illustration of cyclical re-indexing. Node indices are assigned in a circular manner over a fixed range $[0, N)$. A random offset $r$ determines the starting position, and indices wrap around upon reaching the boundary.}
  \label{fig:cyclical_reindex}
\end{figure}

\subsection{Node Cyclic Re-indexing} 
\label{app:reindexing}

In the main text, we describe a cyclic re-indexing scheme to mitigate biases arising from arbitrary node ordering in the Eulerian path serialization. A naive indexing strategy, assigning indices incrementally from $0$ to $n-1$ based on first occurrence, can lead to imbalanced token training, where small-index tokens are over-represented while large-index tokens are under-sampled.

To address this, we adopt the \textit{cyclic re-indexing} scheme introduced in~\cite{zhao2025graphgpt}, which shifts the starting index by a random offset $r$ sampled uniformly from $[0, N)$, where $N$ is a predefined upper bound on node indices (e.g., $N=64$). Specifically, an original index $i$ is mapped to $i' = (i + r) \mod N$. This ensures that all index positions within the range $[0, N)$ are uniformly exposed during training, promoting balanced learning across the token embedding space.

As illustrated in Figure~\ref{fig:cyclical_reindex}, the re-indexing process can be visualized as a circular shift on a ring of $N$ positions. When the index sequence reaches the upper bound $N-1$, it wraps around to $0$, forming a cycle. This circular structure guarantees that no index range is systematically favored, making the model robust to the arbitrary starting point of the Eulerian path.

\begin{figure}[h]
    \centering
    \includegraphics[width=1\linewidth]{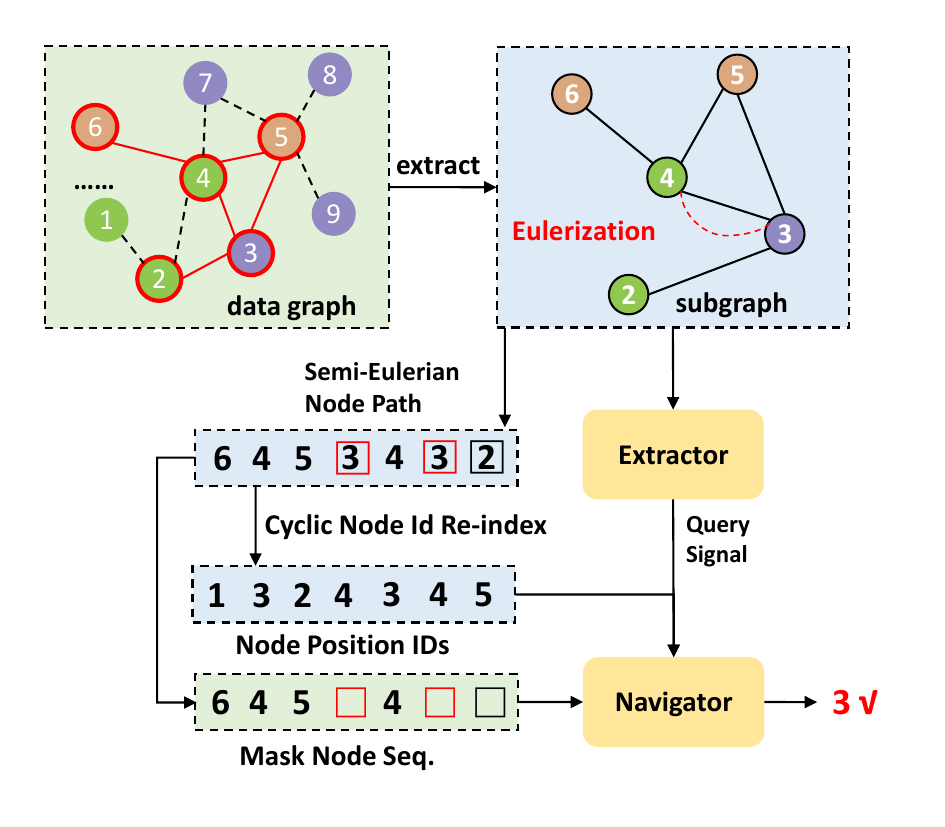}
    \caption{illustration of Training Data Preprocessing.}
    \label{fig:mask_gen_steps}
\end{figure}

\subsection{Data Preprocessing of Training}

This section details the acquisition and processing of training data. As referenced in the main text, our self-supervised training directly utilizes subgraphs sampled from the data graph for masked node prediction, rather than relying on matches obtained by traditional subgraph matching algorithms. This design is motivated by two critical observations:  
\begin{itemize}
    \item \textbf{Computational Intractability}: Enumerating all isomorphic subgraphs in large data graphs is prohibitively expensive. For query graphs with $|V_Q| > 20$, single-match discovery may require seconds-level latency, creating a fundamental barrier for data-driven learning.
    
    \item \textbf{Training Scalability}: As reported in \cite{li2025rsm}, training neural order optimizers with exhaustive matches suffers from severe scalability issues: months-long training for larger queries with limited parallelization gains.
\end{itemize}

To overcome these constraints, we reformulate subgraph matching as intermediate state optimization: (1) Each sampled subgraph serves as a simulated query graph; (2) Masked node prediction approximates the next partial matching states; (3) This abstraction enables full batch parallelism while preserving structural learning objectives.

Figure~\ref{fig:mask_gen_steps} illustrates the complete workflow for a sampled subgraph (nodes with a red border in the data graph):

\begin{enumerate}
    \item Eulerization: Convert subgraph to (semi-)Eulerian path (e.g., $6 \rightarrow 4 \rightarrow 5 \rightarrow 3 \rightarrow 4 \rightarrow 3 \rightarrow 2$).
    
    \item Cyclical Re-indexing: Assign Node Position IDs via $i' = (i + r) \mod N$ with random offset $r$.
    
    \item Masking Strategy:
    \begin{itemize}
        \item Randomly mask $k$ nodes (e.g., $\{2, 3\}$).
        \item Select target node $v_t$ (e.g., $3$).
        \item Replace $v_t$ occurrences with \texttt{[Cls]} tokens.
        \item Replace other masked nodes with \texttt{[Pad]} tokens.
    \end{itemize}
    
    \item Signal Extraction: Feed subgraph to QSExtractor to obtain $h_Q$.
\end{enumerate}

Each training instance is formalized as a tuple: $(\text{Mask Node Sequence},\ \text{Node Position IDs},\ h_{Q},\ v_t)$
where:
\begin{itemize}
    \item $h_Q$: Structural signal from QSExtractor.
    \item $v_t$: Ground-truth node ID for masked position.
\end{itemize}

The GGNavigator learns to predict $v_t$ at \texttt{[Cls]} positions conditioned on $h_Q$ and partial matches. This approach achieves substantially accelerated data generation compared to match-based methods while enabling GPU-parallelized batch processing.

\begin{figure}[h]
    \centering
    \includegraphics[width=1\linewidth]{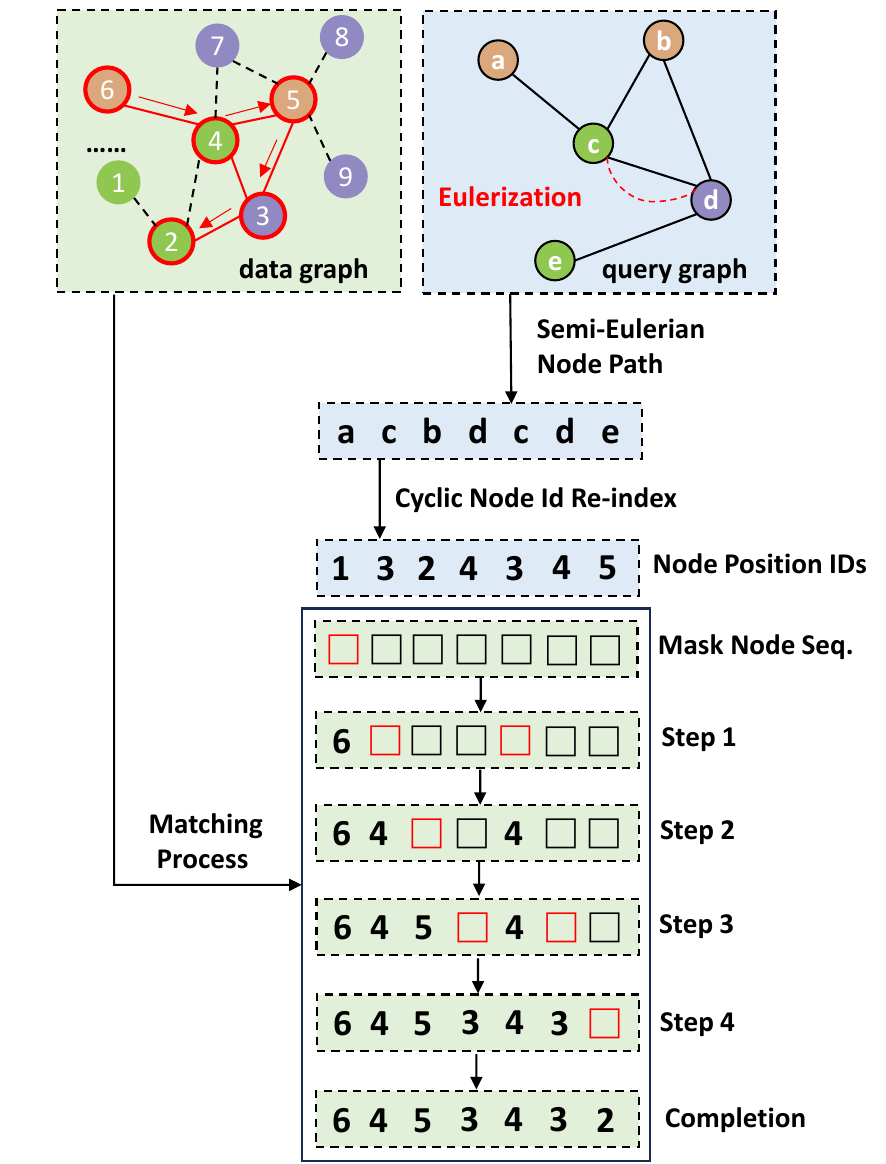}
    \caption{Step-by-step illustration of mask node generation in NeuGN. Red $\square$: class ID (next to match); Black $\square$: padding ID.}
    \label{fig:mask_gen_steps}
\end{figure}

\subsection{Illustration of Mask Node Generation}
\label{app:mask_node_generation}

This section provides a detailed explanation of the generative component in our graph generative navigation framework, specifically, how the prediction from the Mask Node Decoder is integrated with the subgraph matching completion process. As described in the \textit{Euler-guided Mask Node Sequence Generation} section, the (semi-)Eulerian node path guides the generative completion of the Graph Template.

Suppose the matching order of the query graph, determined during the classical ordering phase, is $ a \to c \to b \to d \to e $, and no matching failures occur during the enumeration phase. The step-by-step completion of the Graph Template proceeds as follows (see also Figure~\ref{fig:mask_gen_steps} for an illustrative example):

\begin{enumerate}
    \item \textbf{Initialization}: Through Eulerization, we derive the (semi-)Eulerian node path and construct the initial Graph Template. At this stage, the position corresponding to the first unmatched query node $ a $ is marked with a class ID (denoted by a red $\square$ in the Figure~\ref{fig:mask_gen_steps}), while all other positions are filled with padding IDs (denoted by $\square$).

    \item \textbf{First Matching Step}: The enumerator retrieves the set of local candidate nodes in the data graph for the query node $ a $. The Neural Generative Navigator evaluates these candidates and selects the one with the highest confidence (e.g., node 6) and replaces the class token with its node ID. Subsequently, the class ID is moved to the positions corresponding to the next unmatched query node, $ c $. Since $ c $ appears twice in the (semi-)Eulerian path due to edge duplication during Eulerization, both of its positions are assigned the class IDs.

    \item \textbf{Second Matching Step}: The enumerator now fetches the local candidates for query node $ c $. The navigator selects the most promising candidate (e.g., node 4) and replaces \textbf{all} class tokens with the node 4's ID. The class ID is then advanced to the positions of the next unmatched node, $ b $.

    \item \textbf{Iterative Completion}: This process continues iteratively: at each step, the navigator uses the current partial matching state and structural context to predict the most likely candidate for the node marked by the class token. The selected node ID replaces the class token, and the class token advances to the next relevant positions in the (semi-)Eulerian sequence.
\end{enumerate}

This generation process continues until all padding and class tokens are replaced by actual node IDs from the data graph, resulting in a complete and valid subgraph matching. The integration of neural guidance within this template-filling process enables NeuGN to prioritize high-probability enumeration paths intelligently, significantly reducing unnecessary search.

\section{Details of the Experiments}

This section provides a comprehensive overview of the experimental setup and results. We first describe the six graph datasets used in our evaluation, followed by implementation details for both baseline methods and NeuGN. We then present additional experimental results to support our approach. Furthermore, we report the inference time of our proposed plugin to assess the computational efficiency of the neuro-heuristic framework.

\subsection{Datasets} 

We evaluate our method on six real-world graph datasets spanning diverse domains: social networks (Hamster, LastFM, YouTube), knowledge graphs (NELL), academic collaboration networks (DBLP), and web-based information systems (WikiCS). These datasets exhibit varied topological structures and semantic characteristics, enabling comprehensive evaluation of subgraph matching performance across different graph regimes.

The details of each dataset are as follows:

\begin{itemize}
    \item \textit{Hamster}~\cite{kunegis2013konect} is a social network from the website hamsterster.com, where nodes represent users and edges denote friendship or family relationships. Node classes indicate the type of hamster associated with each user.
    \item \textit{LastFM} \cite{lastfm} captures a social network of LastFM users from Asian countries, collected via the public API. Nodes are users, edges represent mutual follower relationships, and node classes correspond to users' home countries.
    \item \textit{WikiCS} \cite{mernyei2020wikics} is a citation-based graph of Computer Science articles, where nodes represent papers and edges are hyperlinks between them. Node classes denote different subfields within computer science.
    \item \textit{NELL} \cite{carlson2010nell1,yang2016nell2} is derived from the Never-Ending Language Learner system, which continuously extracts facts from web text. The graph consists of entities as nodes and relational facts as edges, with node classes indicating semantic categories.
    \item \textit{DBLP} \cite{yang2012defining} is a co-authorship network where nodes represent authors and edges indicate joint publications. Node classes correspond to four major research areas in computer science, inferred from publication venues.
    \item \textit{YouTube} \cite{yang2012youtube} is a social network of YouTube users, with edges representing friendship relations. Node classes are defined by user-created group memberships, reflecting community interests such as music, gaming, and education.
    
\end{itemize}

\subsection{Implementation Details} 
\paragraph{Implementation of Baselines.}
To evaluate the proposed NeuGN framework, we adopt eight advanced subgraph matching methods for integration, encompassing both traditional heuristic-based and neural-enhanced approaches: VF3~\cite{carletti2017vf3}, QSI~\cite{shang2008qsi}, GQL~\cite{bi2016cfl}, CECI~\cite{bhattarai2019ceci}, CFL~\cite{bi2016cfl}, CaLiG~\cite{yang2023calig}, RLQVO~\cite{wang2022RLQVO}, and RSM~\cite{li2025rsm}. Among them, VF3, QSI, GQL, CECI, CFL, and CaLiG are traditional heuristic-based methods, while RLQVO and RSM are neural-enhanced approaches. 

To ensure a fair comparison, for traditional rule-based methods, we use the implementations integrated in~\cite{zhang2024survey} and enforce single-threaded execution across all methods. For learning-enhanced approaches, we use their publicly available source codes and follow the official hyperparameter settings provided by the original authors.

\paragraph{Implementation Settings.} 
We summarize the key hyperparameters of the proposed NeuGN framework architecture and training process as follows. The model is trained for 1000 epochs with a batch size of 128. In each epoch, the query subgraphs are sampled by performing random walks that visit 5 to 19 distinct nodes starting from each node in the data graph, and the induced subgraph over the visited nodes is extracted as the query. This sampling strategy ensures that larger queries (e.g., 20, 24, and 32 nodes) present in the test set do not appear during training. The token embedding dimension is set to 256. The Transformer decoder consists of 4 layers, each with 8 attention heads, and the hidden dimension of the feed-forward network (FFN) layers is 1024. For the Hamster and LastFM datasets, the decoder is reduced to 2 layers to accommodate their smaller graph scales and lower structural complexity. We optimize the model using the Adam optimizer with a learning rate of $5 \times 10^{-4}$ and employ an exponential learning rate scheduler with a decay factor of $\gamma = 0.999$.

Moreover, our experiments were conducted on a high-performance server equipped with Intel(R) Xeon(R) Gold 6342 CPUs@2.80 GHz, along with NVIDIA A40 GPUs (48GB VRAM). Specifically, the offline training phase was performed using a 4$\times$A40 GPU configuration in parallel, while the online subgraph matching inference was carried out on a single A40 GPU. 

\subsection{Additional Experimental Results}

\textbf{Efficiency Analysis of Batched Neural Inference.} Table~\ref{tab:latency} presents batched inference latency measurements (in milliseconds) on a single NVIDIA A40 GPU with fixed sequence length 64, selected for its representativeness across typical subgraph matching scenarios. All latency values are averaged over 10,000 independent measurement runs to ensure statistical reliability. All reported results incorporate a inference-time optimization: the removal of the softmax layer from the output module. This modification exploits the order, preserving property of linear scoring, specifically, $\text{argsort}(\mathbf{W}h_{[\text{Cls}]}) \equiv \text{argsort}(\text{softmax}(\mathbf{W}h_{[\text{Cls}]}))$, which ensures identical candidate rankings while eliminating the computational overhead of exponentiation and normalization. The results demonstrate practical operational efficiency: after amortization of latency through batch processing (BS = 16), the inference of the per query is completed in a submillisecond time for most datasets (0.06 to 0.75 ms). Even for the YouTube graph (1.13M nodes), latency remains well below 3ms (2.43ms), comfortably satisfying real-time constraints for applications. This efficiency stems from NeuGN's GPU-accelerated design, where batching distributes fixed computational costs across multiple queries. The observed latency progression, increasing predictably with graph node vocabulary size, aligns with architectural expectations, confirming measurement validity.

\begin{table}[]
\small
\setlength{\tabcolsep}{0.95mm}
\begin{tabular}{lcccccc}
\toprule
 Dataset& 
  \multicolumn{1}{c}{Hamster} &
  \multicolumn{1}{c}{LastFM} &
  \multicolumn{1}{c}{WikiCS} &
  \multicolumn{1}{c}{NELL} &
  \multicolumn{1}{c}{DBLP} &
  \multicolumn{1}{c}{YouTube} \\
  \midrule
BS=1      & 0.96 & 0.96 & 1.52 & 1.55 & 2.03  & 3.71  \\
BS=4     & 1.01 & 1.04 & 1.64 & 2.04 & 4.33  & 11.12  \\
BS=16     & 1.03 & 1.24 & 1.99 & 3.84 & 12.05 & 38.91 \\ \midrule
PQ(B=4) & 0.25& 0.26&0.41&0.51&1.08&2.78\\
PQ(B=16) & 0.06 & 0.08 & 0.12 & 0.24 & 0.75  & 2.43\\ 
\bottomrule
\end{tabular}
\caption{Batched inference latency (ms) of GGNavigator with sequence length=64. Total latency for batch sizes and per-query amortized latency.}
\label{tab:latency}
\end{table}
\textbf{Training Efficiency Analysis.} 
Table~\ref{tab:training_time} presents the per-epoch training time across datasets using our full-node sampling strategy, executed on a parallelized 4$\times$NVIDIA A40 GPU configuration. This design choice ensures comprehensive coverage of local structural patterns by treating each node as an anchor point for query subgraph generation, thereby preventing any structural motifs from being overlooked during training. The training time exhibits a near-linear relationship with graph size, increasing from 0.02 minutes on Hamster (2.4K nodes) to 19.30 minutes on YouTube (1.13M nodes). The observed time complexity aligns with the theoretical $O(|V|)$ expectation of our sampling methodology, where $|V|$ represents the number of vertices in the data graph.

\begin{table}[H]
\small
\setlength{\tabcolsep}{0.9mm}
\begin{tabular}{@{}lcccccc@{}}
\toprule
Dataset  & Hamster & LastFM & WikiCS & NELL & DBLP & YouTube \\ 
\midrule
Time (min) & 0.02    & 0.05   & 0.11   & 0.38 & 3.09 & 19.30   \\ 
\bottomrule
\end{tabular}
\caption{Per-epoch training time (minutes) across datasets}
\label{tab:training_time}
\end{table}

\textbf{Details and Additional Results of Early Convergence Acceleration (RQ5).}

This section provides supplementary details to the early convergence experiment in the main text. All NeuGN inference operations were executed on a single NVIDIA A40 GPU with the following constrained configuration: batch size fixed at 4, navigation depth set to 16, and inference iterations limited to 24 to avoid computational overhead. Extended results demonstrating acceleration performance are presented in Table~\ref{tab:add}.

\begin{table*}[h]
\centering
\begin{tabular}{lrrrrrr}
\toprule
 & \multicolumn{1}{c}{Hamster} & \multicolumn{1}{c}{LastFM} & \multicolumn{1}{c}{WikiCS} & \multicolumn{1}{c}{NELL} & \multicolumn{1}{c}{DBLP} & \multicolumn{1}{c}{YouTube} \\ \midrule
GQL & 4.29E+05 & 8.22E+05 & 3.06E+05 & 4.58E+05 & 8.63E+05 & 1.83E+05 \\
+NeuGN & 5.02E+05 & 1.14E+06 & 4.67E+05 & 5.98E+05 & 9.02E+05 & 2.21E+05 \\
Improv. & \textbf{17.0\%} & \textbf{39.0\%} & \textbf{52.4\%} & \textbf{30.5\%} & \textbf{4.5\%} & \textbf{20.7\%} \\ \midrule
CFL & 5.42E+06 & 6.50E+06 & 3.41E+06 & 6.02E+06 & 1.19E+07 & 4.82E+06 \\
+NeuGN & 5.70E+06 & 7.05E+06 & 5.30E+06 & 7.15E+06 & 1.22E+07 & 5.12E+06 \\
Improv. & \textbf{5.2\%} & \textbf{8.5\%} & \textbf{55.6\%} & \textbf{18.7\%} & \textbf{2.1\%} & \textbf{6.2\%} \\ \midrule
CECI & 5.15E+06 & 5.04E+06 & 4.93E+06 & 4.79E+06 & 7.47E+06 & 5.07E+06 \\
+NeuGN & 5.62E+06 & 5.94E+06 & 6.79E+06 & 5.76E+06 & 7.74E+06 & 5.59E+06 \\
Improv. & \textbf{9.2\%} & \textbf{17.8\%} & \textbf{37.7\%} & \textbf{20.2\%} & \textbf{3.6\%} & \textbf{10.3\%} \\ \midrule
CaLiG & 6.58E+06 & 5.54E+06 & 4.43E+06 & 4.46E+06 & 9.89E+06 & 2.22E+06 \\
+NeuGN & 6.78E+06 & 6.28E+06 & 8.50E+06 & 6.09E+06 & 1.01E+07 & 2.48E+06 \\
Improv. & \textbf{3.0\%} & \textbf{13.3\%} & \textbf{91.6\%} & \textbf{36.5\%} & \textbf{2.0\%} & \textbf{11.6\%} \\ \bottomrule
\end{tabular}
\caption{Additional Results of RQ5.}
\label{tab:add}
\end{table*}

\section{Discussion}
\subsection{Limitation} 
NeuGN improves subgraph matching only through permutational navigation: it reorders candidate vertices but never prunes them, thereby preserving completeness at the cost of leaving the search tree size essentially unchanged. Although a neural classifier that eliminates hopeless branches could in theory yield greater speed-ups, integrating such pruning remains impractical because any false-negative decision would break completeness and undermine the exact-matching guarantee.

\subsection{Future Work}
\textbf{ Graph Tokenization for Web-Scale Deployment.} 
To effectively alleviate the vocabulary explosion problem encountered in web-scale graphs ($>100$M nodes), we plan to explore hierarchical tokenization strategies for NeuGN based on METIS partitioning~\cite{karypis1998metis,karypis1998multimetis}, using dual-token identifiers $[\text{PartitionID}, \text{LocalNodeID}]$. This approach compresses the node vocabulary size from $O(|V|)$ to $O(\sqrt{|V|})$, significantly reducing model memory footprint, while aiming to preserve essential topological context within and across partitions.\\
\textbf{Semantic-aware Matching for Complex Structures.} 
Classical subgraph matching algorithms are limited in handling high-dimensional semantic graphs (e.g., text-attributed academic concept networks) due to their reliance on rigid, rule-based heuristics. In contrast, NeuGN’s inherent semantic awareness opens a pathway toward future cross-modal subgraph matching, aligning topological patterns with unstructured text or visual attributes.

\end{document}